\documentclass[a4paper,conference]{IEEEtran}
\ifCLASSINFOpdf
\else
\fi
\usepackage{hyperref}

\usepackage{xcolor}
\usepackage{amsfonts, bm}       
\usepackage{algorithm}
\usepackage{algorithmic} 
\usepackage{graphicx}
\usepackage{booktabs}       

\begin{document}
%
\title{ARCADe: A Rapid Continual Anomaly Detector}

\author{\IEEEauthorblockN{Ahmed Frikha}
\IEEEauthorblockA{University of Munich\\
Siemens AG, Corporate Technology\\
Munich, Germany\\
ahmed.frikha@siemens.com}
\and
\IEEEauthorblockN{Denis Krompa{\ss}}
\IEEEauthorblockA{Siemens AG, Corporate Technology\\
Munich, Germany\\
denis.krompass@siemens.com}
\and
\IEEEauthorblockN{Volker Tresp}
\IEEEauthorblockA{University of Munich\\
Siemens AG, Corporate Technology\\
Munich, Germany\\
volker.tresp@siemens.com}}

\maketitle

\begin{abstract}
Although continual learning and anomaly detection have separately been well-studied in previous works, their intersection remains rather unexplored. The present work addresses a learning scenario where a model has to incrementally learn a sequence of anomaly detection tasks, i.e. tasks from which only examples from the normal (majority) class are available for training. We define this novel learning problem of continual anomaly detection (CAD) and formulate it as a meta-learning problem. Moreover, we propose \emph{A Rapid Continual Anomaly Detector (ARCADe)}, an approach to train neural networks to be robust against the major challenges of this new learning problem, namely catastrophic forgetting and overfitting to the majority class. The results of our experiments on three datasets show that, in the CAD problem setting, ARCADe substantially outperforms baselines from the continual learning and anomaly detection literature. Finally, we provide deeper insights into the learning strategy yielded by the proposed meta-learning algorithm. 
\end{abstract}


%
\IEEEpeerreviewmaketitle

\section{Introduction} \label{intro}
Humans can continually learn new tasks without corrupting their previously acquired abilities. Neural networks, however, tend to overwrite older knowledge and therefore fail at incrementally learning new tasks. This is called the catastrophic forgetting problem \cite{french1999catastrophic, goodrich2015neuron, kirkpatrick2017overcoming}. In fact, most deep learning achievements have been realized in the offline supervised single-task or multi-task learning \cite{caruana1997multitask} settings, where the availability of independent and identically distributed (i.i.d.) data can be assumed. Building intelligent agents that are able to incrementally acquire new capabilities while preserving the previously learned ones remains an old and long-standing goal in machine learning research.     \\

Several approaches have been developed to enable continual learning, e.g. by alleviating interference between the sequentially learned tasks, \cite{zenke2017continual, kirkpatrick2017overcoming, lee2017overcoming} and/or encouraging knowledge transfer between them \cite{lopez2017gradient, chaudhry2018riemannian, riemer2018learning, javed2019meta, beaulieu2020learning}. While most of the previous works addressed the continual learning problem with neatly class-balanced classification tasks, many real-world applications exhibit extreme class-imbalance, e.g. in anomaly detection \cite{chandola2009anomaly} problems. For example, in industrial manufacturing, of all produced parts, only a few per million are faulty. And since the products and/or machines in the plant are continuously changing, building a central anomaly detector that incrementally improves by learning new anomaly detection tasks would relax this cold-start problem.\\

To the best of our knowledge, continual learning with class-imbalanced data has only been addressed in \cite{aljundi2019task, wang2018systematic}. Hereby the authors assume, however, access to examples from all classes, including the minority class. In the anomaly detection literature \cite{chandola2009anomaly} most works address the unsupervised anomaly detection problem, where only examples from the majority (normal) class are available for training an anomaly detector. Learning a binary classifier using data samples from only one of its classes (usually the majority class) is referred to as One-Class Classification (OCC) \cite{moya1993one, khan2014one}. Our work addresses the novel and unexplored problem of \emph{Continual Anomaly Detection (CAD)}, where different binary classification tasks have to be learned sequentially by using only examples from their respective majority classes for training. We also refer to this problem as \emph{Continual One-Class Classification (ContOCC)}. In particular, we propose an approach that relies on meta-learning \cite{schmidhuber1987evolutionary} to yield a parameter initialization that resists to the main challenges of CAD, namely catastrophic forgetting and overfitting to the majority class. Several state-of-the-art works introduced meta-learning algorithms to tackle continual learning problems \cite{riemer2018learning, javed2019meta, spigler2019meta, zhang2019variational, beaulieu2020learning}. Hereby, however, only class-balanced classification tasks were considered. \\

Our contribution in this work is threefold: Firstly, we introduce and define the novel and relevant CAD problem. Secondly, we propose a first, strong and model-agnostic approach to tackle it. Thirdly, we successfully validate our approach on three datasets, where we substantially outperform continual learning and anomaly detection baseline methods.

\section{The Continual Anomaly Detection (CAD) Problem} \label{CAD-problem}

The goal of \emph{Continual Anomaly Detection (CAD)} or \emph{Continual One-Class Classification (ContOCC)} is to sequentially learn multiple OCC tasks without forgetting previously learned one. More precisely, the target model should be able to sequentially learn binary classification tasks by using only examples from their respective normal classes for training, and then achieve a high performance in distinguishing between both classes of each of the learned tasks, when faced with unseen datapoints. The CAD problem is a prototype for a practical use case where a central anomaly detector for multiple applications is needed and new applications become available gradually in time. In this section, we first discuss the unique challenges of the CAD learning scenario. Subsequently, we present a problem formulation for CAD. Finally, we introduce the meta-learning optimization technique, upon which our approach builds to tackle CAD.

\subsection{Unique Challenges} \label{CAD-challenges}
In order to perform CAD, approximating \emph{one} decision boundary that encompasses \emph{all} the (normal) majority classes of the observed tasks is necessary. In fact, the examples belonging to the normal class of \emph{any} observed task should be mapped inside the normal class boundary, and therefore classified as normal. Learning such a decision boundary can be especially challenging due to two inherent problems of neural networks: catastrophic forgetting and overfitting to the majority class, i.e. predicting the normal class label for any input. On the one hand, each model update that we perform using examples from a new task shifts our decision boundary away from the normal class of previously learned tasks, resulting in a poorer classification performance on the latter (catastrophic forgetting). On the other hand, since the model is only exposed to (normal) majority class examples, the decision boundary tends to  over-generalize and classify any input as normal. This way the model overfits to the normal class and anomalies would not be detected.

\subsection{Problem Formulation} \label{problem-formulation}
We define a CAD task-sequence $S=\{T_1, ..., T_n\}$ as an ordered sequence of OCC tasks $T_i$. To learn $S$, the classification model is trained on the tasks included in it, one after another. Due to the sequential exposure to tasks, the model is trained with non i.i.d. samples. This setting is commonly used in class-balanced continual learning to define non-stationary conditions. It is also called \emph{locally i.i.d.} \cite{lopez2017gradient, riemer2018learning}, since the model is exposed to a sequence of stationary distributions, defined by the tasks $T_i$. In contrast, offline single-task and multi-task learning assume that a fixed training dataset is available at all points in time. We note that for an OCC task $T_i$, the training set $T_{i}^{tr}$ and the validation set $T_{i}^{val}$ have different data distributions, since $T_{i}^{tr}$ includes only examples from one class and $T_{i}^{val}$ is class-balanced. \\

In the following we formulate the CAD problem as a meta-learning problem. We consider separate sets of task-sequences for meta-training ($D_{tr}$), meta-validation ($D_{val}$) and meta-testing ($D_{test}$). Hereby all the tasks in these sequences belong to the same domain, i.e. come from a task distribution $p(T)$. To prevent leakage between $D_{tr}$, $D_{val}$ and $D_{test}$, these sets of data must have mutually exclusive classes, i.e. none of the classes building the tasks $T$ included in $D_{tr}$ is used to build a task in $D_{val}$ or $D_{test}$ and vice versa. \\

Each sequence in $D_{tr}$, $D_{val}$ and $D_{test}$ is composed of a training and a test set. Let $S_{tr}= \{S_{tr}^{tr}, S_{tr}^{val}\}$ denote a meta-training task-sequence from $D_{tr}$, where $S_{tr}^{tr}=\{T^{tr}_{1}, ..., T^{tr}_{n}\}$ is a sequence of the training sets of the tasks composing $S$, and $S_{tr}^{val}=\{T^{val}_{1}, ..., T^{val}_{n}\}$ is a sequence of their validation sets. Following the terminology introduced in \cite{beaulieu2020learning} to formulate meta-learning problems, we refer to $S_{tr}^{tr}$ as a \emph{meta-training training} task-sequence and $S_{tr}^{val}$ as a \emph{meta-training validation} task-sequence. We call the set of all meta-training training (validation) sequences the meta-training training (validation) set $D_{tr}^{tr}$ ($D_{tr}^{val}$). We note that the sequences in $D_{tr}^{tr}$ include examples from only the majority class of each task, while the sequences in $D_{tr}^{val}$ contain disjoint class-balanced sets of data from each task. The same holds for the meta-validation and meta-testing sets $D_{val}$ and $D_{test}$. \\

We aim to find an algorithm that, by using $D_{tr}$, yields a learning strategy that enables a classification model to sequentially learn anomaly detection tasks without (or with minimal) forgetting. Applying this learning strategy to a random task-sequence from $D_{test}^{tr}$ would then provide a model that has high performance on $D_{test}^{val}$, hence performing CAD. In this work the learning strategy yielded by the proposed meta-learning algorithm consists in a model initialization and a learning rate for each model parameter, which are suitable to perform CAD. Starting from the meta-learned model initialization, taking few gradient descent steps with the meta-learned learning rates to learn each of the OCC tasks in a sequence $S$ leads to a proficient anomaly detector on all tasks.

\subsection{Continual Learning via Meta-Learning a Parameter Initialization} \label{parameter-init}
The proposed meta-learning approach to tackle the CAD problem learns a model initialization and parameter-specific learning rates by building upon a bi-level optimization scheme. In this section we explain this optimization mechanism which was introduced in the MAML algorithm \cite{finn2017model} to address the few-shot learning problem \cite{vinyals2016matching, wang2019few}. Since then this optimization scheme was used by multiple meta-learning algorithms to address several problems, e.g. few-shot learning \cite{finn2017model, li2017meta, lee2019meta, raghu2019rapid}, few-shot one-class classification \cite{frikha2020fewshot}, resisting to adversarial examples \cite{yin2018adversarial} and continual learning \cite{javed2019meta, spigler2019meta, beaulieu2020learning}. \\

Let $\theta$ denote the set of model parameters. The aforementioned bi-level optimization mechanism aims to optimize these model parameters to be easily adaptable to unseen tasks $T_i$ which have certain characteristics, e.g. few-shot learning tasks, anomaly detection tasks or continual learning tasks. After adaptation to a task $T_i$, e.g. by taking few gradient steps using its training set, the adapted parameters $\theta^{'}_{i}$ yield high performance on a held-out test set of the same task. In that sense the meta-learned model parameters $\theta$ can be viewed as a parameter initialization that enables quick learning of unseen tasks. The meta-learned parameter initialization represents an inductive bias that facilitates learning tasks with certain characteristics. \\

To find such a model initialization, a model is explicitly trained for quick adaptation using a set of meta-training tasks. Hereby, these tasks belong to the same domain and have the same characteristics as the test tasks, e.g. if the unseen test tasks are expected to have only few examples, the meta-training tasks should be few-shot learning tasks \cite{finn2017model}. In each meta-training iteration, two operations are performed for each task, parameter adaptation and evaluation. Adapting the model initialization $\theta$ to a task $T_i$ is done by taking few gradient descent steps using its training set $T_{i}^{tr}$, yielding a task-specific model $\theta^{'}_{i}$. The evaluation of the task-specific model uses the task's validation set $T_{i}^{val}$. The resulting loss $L^{val}_{T_{i}}(f_{\theta^{'}_{i}})$ is used to update the initialization $\theta$ as shown in Equation \ref{eq:outer-update}, where $\beta$ is the learning rate used for this update.

\begin{equation} 
\label{eq:outer-update}
  \theta \gets \theta - \beta \nabla_{\theta} \sum_{T_{i}\sim p(T)} L^{val}_{T_{i}}(f_{\theta^{'}_{i}}).
\end{equation}

For a model initialization to be suitable for continual learning, i.e. to inhibit catastrophic forgetting, each meta-training and meta-testing task is built as a sequence of classification tasks \cite{spigler2019meta, javed2019meta}. Hereby, adapting the parameter initialization to a task-sequence consists in taking a few gradient descent steps on the tasks included in it, sequentially. The parameter initialization is then updated as shown in Equation \ref{eq:outer-update}, where $L^{val}_{T_{i}}(f_{\theta^{'}_{i}})$ is the sum of the losses computed on the validation set of each task in the task-sequence.

\section{Related Work} \label{related-works}
The present work addresses the Continual Anomaly Detection (CAD) problem, which represents the intersection of the continual learning and anomaly detection problems. To the best of our knowledge no prior works addressed the CAD problem. Therefore, in this section we review related continual learning and anomaly detection work separately.

\subsection{Continual Learning}
Several Continual learning (CL) approaches inhibit catastrophic forgetting by retaining past knowledge. This can be done by increasing the model capacity \cite{rusu2016progressive} or by regularizing the parameter updates \cite{kirkpatrick2017overcoming, zenke2017continual, lee2017overcoming}. Another category of CL methods relies on replaying previous experiences, e.g. datapoints, by interleaving them between new experiences \cite{schaul2015prioritized, lopez2017gradient, riemer2018learning}. Recent works developed meta-learning based approaches to tackle CL \cite{riemer2018learning, zhang2019variational, he2019task, javed2019meta, spigler2019meta, beaulieu2020learning}. In \cite{riemer2018learning}, a method that maximizes transfer and minimizes interference between the sequentially learned tasks is developed by combining the meta-learning algorithm Reptile \cite{nichol2018reptile} with a reservoir sampling. The CL approach proposed in \cite{zhang2019variational} learns and continuously adapts class prototypes, by building upon the meta-learning algorithm ProtoNets \cite{snell2017prototypical}. \\

Using the bi-level optimization scheme introduced in Section \ref{parameter-init}, methods were developed to meta-learn a parameter initialization that inhibits catastrophic forgetting \cite{javed2019meta, spigler2019meta, beaulieu2020learning}. Here, it is possible to learn an initialization for all model parameters \cite{spigler2019meta} or learn an embedding network and an initialization for only the classifier network \cite{javed2019meta, beaulieu2020learning}. In \cite{beaulieu2020learning}, a separate network is additionally trained to perform a task-specific feature weighting by modulating the output of the embedding network. The aforementioned works address CL by assuming that the classification tasks, which have to be learned, are \emph{class-balanced}. The absence of any mechanism to cope with the extreme setting, where all the tasks are OCC tasks as in CAD, makes these approaches prone to overfitting to the majority class. In contrast, our approach inhibits this undesired phenomenon besides reducing catastrophic forgetting. We compare to the meta-learning based continual learning algorithm SeqFOMAML \cite{spigler2019meta} in our experiments and show that it overfits to the majority class in the CAD problem setting.

\subsection{Anomaly Detection and One-Class Classification}
Typical anomaly detection (AD) approaches use SVMs to detect anomalous examples \cite{scholkopf2001estimating, tax2004support}, i.e. examples that do not belong to the normal class. When faced with high-dimensional data, e.g. images, feature extractors are used to embed the data into a lower-dimensional space before they are fed to the SVM-based classifier \cite{Xu_2015, andrews2016transfer, erfani2016high}. End-to-end deep learning methods were also proposed to tackle AD, by jointly training a feature extractor and a one-class classifier \cite{ruff2018deep} or by using the reconstruction loss of autoencoders \cite{hinton2006reducing} to distinguish anomalies \cite{hawkins2002outlier, an2015variational, chen2017outlier}. GAN-based \cite{goodfellow2014generative} approaches were also used for AD \cite{Ravanbakhsh_2017, schlegl2017unsupervised, Sabokrou_2018}. Recently, an episodic data sampling strategy was proposed to adapt various class-balanced meta-learning algorithms to the AD setting \cite{frikha2020fewshot}. Hereby, the bi-level optimization mechanism explained in Section \ref{parameter-init} is used to find a model (initialization) that enables few-shot AD, i.e. learning a classification task by using only few examples from only its normal class.  \\

All the aforementioned approaches yield a classification model that can detect the anomalies of a \emph{single} AD task. In fact, they do not incorporate any feature to promote learning multiple tasks sequentially or inhibit catastrophic forgetting, which makes them unsuitable for the CAD problem setting. We propose a method that enables a model to sequentially learn \emph{multiple} AD tasks with only minimal forgetting. In our experiments, we compare to the meta-learning algorithm OC-MAML \cite{frikha2020fewshot}, which yields an initialization tailored for learning AD tasks. Our results (Section \ref{results}) show that it fails at sequentially learning several tasks.

\section{Approach: A Rapid Continual Anomaly Detector (ARCADe)} \label{approach}
This work introduces \emph{A Rapid Continual Anomaly Detector (ARCADe)}, a meta-learning algorithm designed to tackle the Continual Anomaly Detection (CAD) problem (Section \ref{CAD-problem}). ARCADe builds upon the bi-level optimization scheme introduced in Section \ref{parameter-init}. Since meta-learning algorithms that use this optimization mechanism have been shown to be universal learning algorithm approximators \cite{finn2017metalearning}, ARCADe should be able to approximate a learning algorithm tailored for the CAD problem. In this section, we first present ARCADe using the CAD problem formulation from Section \ref{problem-formulation}. Subsequently, we explain the intuition behind meta-learning parameter-specific learning rates. Finally, we destinguish between two variants of ARCADe.

\subsection{Algorithm}
Our algorithm uses the meta-training set $D_{tr}$ to learn an initialization $\theta$ as well as a learning rate $\alpha$ for each model parameter, as done in \cite{li2017meta} to address the few-shot learning problem. Starting from this meta-learned initialization, learning a sequence $S_{test}$ of unseen OCC tasks (by taking few gradient descent steps) using the meta-learned learning rates yields a model that has a high performance on all tasks included in $S_{test}$. The meta-training procedure of ARCADe is presented in Algorithm \ref{meta-training-algo}.

\begin{algorithm}
\caption{ARCADe Meta-training Procedure}
\label{meta-training-algo}
\begin{algorithmic}[1]
\REQUIRE $D_{tr}$: Set of meta-training task-sequences 
\REQUIRE $\beta$: Learning rate for the meta-update
\REQUIRE $K$: Adaptation set size
\STATE Randomly initialize model parameters $\bm{\theta}$ and parameter-specific learning rates $\bm{\alpha}$
\WHILE{not done}
  \STATE Sample a batch of task-sequences $S_i$ from $D_{tr}$
  \STATE Initialize meta-learning loss $L_{meta}=0$ 
  \FOR{\textbf{each} sampled $S_i$}
    \STATE Initialize sequence adaptation loss $L_s = 0$
    \STATE Initialize $\bm{\theta^{'}_{i,0}}$ = $\bm{\theta}$ ($\bm{\theta^{'}_{i,0}}$ = $\bm{\theta_{head}}$ if ARCADe-H)
    \FOR{$T_j$ in $S_i$ with $j$ in $\{1, ..., J= \emph{length($S_i$)}\}$} 
      \STATE Compute adapted parameters using $K$ (normal) majority class examples from $T^{tr}_{j}$:\\
      $\bm{\theta^{'}_{i,j}} = \bm{\theta^{'}_{i,j-1}} - \bm{\alpha} \circ \nabla_{\bm{\theta^{'}_{i,j-1}}} L_{T^{tr}_{j}}(f_{\bm{\theta^{'}_{i,j-1}}})$
      \STATE Compute $L_{T^{val}_{j}}(f_{\bm{\theta^{'}_{i,j}}})$ with the current adapted parameters $\bm{\theta^{'}_{i,j}}$ on the class-balanced val set $T^{val}_{j}$
      \STATE $L_s = L_s + L_{T^{val}_{j}}(f_{\bm{\theta^{'}_{i,j}}})$
    \ENDFOR
    \FOR{$T_j$ in $S_i$}
      \STATE Compute loss $L_{T^{val}_{j}}(f_{\bm{\theta^{'}_{i,J}}})$ with the final adapted parameters $\bm{\theta^{'}_{i,J}}$ on the val set $T^{val}_{j}$
      \STATE $L_s = L_s + L_{T^{val}_{j}}(f_{\bm{\theta^{'}_{i,J}}})$
    \ENDFOR
    \STATE $L_{meta} = L_{meta} + L_s$
  \ENDFOR
  \STATE Update $(\bm{\theta},\bm{\alpha})$: $(\bm{\theta},\bm{\alpha}) \gets (\bm{\theta},\bm{\alpha}) - \beta \nabla_{(\bm{\theta},\bm{\alpha})} L_{meta}$
\ENDWHILE
\RETURN Meta-learned parameters $\bm{\theta}$ and learning rates $\bm{\alpha}$
\end{algorithmic}
\end{algorithm}

In each meta-training iteration of ARCADe a batch of task-sequences is randomly sampled from $D_{tr}$. The current parameter initialization $\bm{\theta}$ is adapted to each sequence $S_i$ by taking one (or more) gradient step(s) on the training sets $T^{tr}_j$ of the tasks included in $S_i$ sequentially. Hereby the gradient descent steps are performed using the current parameter-specific learning rates $\bm\alpha$. We note that in Algorithm \ref{meta-training-algo} only one gradient descent update is performed (Operation 9) for simplicity of notation. Extending it to multiple updates is straightforward. We use the binary cross-entropy loss for all loss functions mentioned in Algorithm \ref{meta-training-algo}. \\

In the CAD problem setting (Section \ref{problem-formulation}), we consider anomaly detection tasks (or OCC tasks), i.e. each task $T_j$ includes a training set $T_j^{tr}$ with only majority class examples and a class-balanced validation set $T_j^{val}$. For each task $T_j$ we compute the loss on the class-balanced held-out validation set $T^{val}_j$ twice. The first time (Operation 10) is done directly after learning $T_j$ by using the adapted model $\bm{\theta^{'}_{i,j}}$. This ensures a high model performance on the task immediately after it is learned. The second time (Operation 14) is conducted after learning all the tasks in $S_i$, i.e. using the final model adapted to that sequence $\bm{\theta^{'}_{i,J}}$. This maximizes the last model's performance on all the tasks in the sequence, hence minimizing catastrophic forgetting. These two losses are computed for \emph{each} task in $S_i$ and added to the sequence adaptation loss $L_s$. The model initialization and learning rates are updated in each meta-training iteration by minimizing $L_{meta}$ which is the sum of the adaptation losses $L_s$ of each sampled task-sequence $S_i$ (Operation 19). In that sense, we can say that ARCADe explicitly optimizes for having a high performance on all tasks contained in a sequence, immediately after learning them and after having learned them all sequentially, while using only examples from their majority class. \\

In order to ensure that the model has a high performance on a task $T_j$ at all points in time after learning it, one could compute the loss on its validation set $T^{val}_j$ after learning \emph{each} task $T_k$ subsequent to $T_j$ and add it to $L_s$. Here the loss would be computed using the current model parameters $\bm{\theta^{'}_{i,k}}$ after learning a task $T_k$. Doing this would minimize forgetting task $T_j$ in all points in time while incrementally learning new tasks ($T_k$). However, in this case, the computational cost for computing $L_s$ would increase exponentially with the length of the task-sequence, which does not scale for long task-sequences. Instead we approximate this additional optimization objective by adding to $L_s$ the validation loss of \emph{one} randomly sampled previous task $T_j$, every time a new task $T_k$ in the sequence is learned. We note that this cannot be performed for the first task in the sequence, since it has no previous tasks. Even though we compute these additional loss terms and use them for our experimental evaluation, we do not mention them in Algorithm \ref{meta-training-algo} for simplicity of notation.  \\

Once meta-training is done, the best performing initialization and learning rates are used to learn task-sequences from the meta-testing set $D_{test}$. Here, the model initialization is sequentially adapted to the tasks from the test task-sequence using their training sets and the meta-learned learning rates, as done during meta-training (Operations 8 and 9 in Algorithm \ref{meta-training-algo}). Thereafter the adapted model is evaluated on the class-balanced validation sets of these tasks, as done in meta-training (Operations 13 and 14 in Algorithm \ref{meta-training-algo}). We note that the selection of the best performing model initialization and learning rates is done by conducting validation episodes (adaptation and evaluation) using the task-sequences from the meta-validation set $D_{val}$, throughout meta-training.

\subsection{Meta-Learning Parameter-Specific Learning Rates}
In the following, we explain the intuition behind additionally meta-learning parameter-specific learning rates to tackle the CAD problem and not only the model initialization as it was done in \cite{spigler2019meta}, \cite{javed2019meta} and \cite{beaulieu2020learning} in the class-balanced continual learning setting. We hypothesize that meta-learning parameter-specific learning rate enables the optimization algorithm to identify the parameters that are responsible for overfitting to the majority class and/or for catastrophic forgetting, and reduce their learning rates. Our results (Section \ref{results}) confirm our intuition and show that additionally meta-learning parameter-specific learning rates leads to a more effective inductive bias for the CAD problem. \\

Before performing the adaptation updates (Operation 9 in Algorithm \ref{meta-training-algo}), we clip the learning rates to have values between $0$ and $1$. We do this to prevent them from having negative values, which would lead to taking gradient ascent steps on the task adaptation loss $L_{T^{tr}_{j}}$. The meta-update (Operation 19 in Algorithm \ref{meta-training-algo}) can indeed update the learning rates to have negative values since performing gradient ascent on the one-class training set of a task prevents overfitting to that class (by increasing the loss on that class). The lower overfitting to the majority class leads to a lower loss on the class-balanced validation set ($L_{T^{val}_{j}}(f_{\bm{\theta^{'}_{i,j}}})$), which results in a lower $L_{meta}$. By clipping the negative learning rates to $0$, we ensure that the corresponding parameter (responsible to overfitting to the majority class) is not updated during task-adaptation. It is considered as a task-agnostic parameter and is used as-is for all tasks, as opposed to other parameters which are updated to task-specific values. To speed-up meta-training, it is possible to conduct the first $n$ meta-training iterations with constant learning rates, before meta-learning them along with the initialization (Operation 13 in Algorithm \ref{meta-training-algo}).

\subsection{Variants of ARCADe}
We distinguish two variants of ARCADe: ARCADe-M, which we introduced up to now, meta-learns an initialization and a learning rate for \emph{all} model parameters, and ARCADe-H, which does the same but only for the parameters of the classification head, i.e. the output layer. For the parameters of the backbone layers, ARCADe-H does not learn an initialization but rather task-agnostic end values, which do not have to be updated depending on the task-sequence that has to be learned. When learning tasks sequentially ARCADe-H updates only the parameters of the output layer with their corresponding meta-learned learning rates. The only difference in the meta-learning procedure can be seen in Operation 7 from \mbox{Algorithm \ref{meta-training-algo}}. Meta-learning approaches that adapt only the classification head to learn unseen tasks were proposed in \cite{raghu2019rapid} and \cite{javed2019meta} to address the few-shot learning and the class-balanced continual learning problems, respectively. 

\section{Experimental Evaluation} \label{experiments}
We conduct experiments \footnote{Our code is made public under: \url{https://github.com/AhmedFrikha/ARCADe-A-Rapid-Continual-Anomaly-Detector}} in an attempt to answer the following key questions: $(1)$ Can the proposed meta-learning algorithm cope with the challenges of the CAD problem, i.e. catastrophic forgetting and overfitting to the majority class, and how do its two variants, ARCADe-M and ARCADe-H, compare to each other? $(2)$ How do previous meta-learning approaches for anomaly detection and class-balanced continual learning perform in the CAD setting? $(3)$ Does meta-learning a learning rate for each parameter, besides the initialization, boost performance in a CAD context? $(4)$ If yes, does the distribution of the meta-learned learning rates follow a pattern across datasets?

\subsection{Baselines and Datasets}
We evaluate the two variants of the proposed meta-learning Algorithm (ARCADe-M and ARCADe-H) on three different datasets which range from  grey-scale images of letters to more challenging RGB natural images (Question $1$). Besides we compare ARCADe to OC-MAML \cite{frikha2020fewshot} and SeqFOMAML \cite{spigler2019meta}, which meta-learn model initializations that are tailored for anomaly detection and continual learning, respectively (Questions $2$). We use the same evaluation procedure for ARCADe and the baselines: Task-sequences are sampled from the meta-testing set $D_{test}$ and their tasks are learned sequentially using gradient descent. For a fairer comparison, we adapt SeqFOMAML to the anomaly detection scenario by using anomaly detection tasks for its meta-training. Note that SeqFOMAML samples the same number of examples from each class during the adaptation phase of its meta-training, i.e. it uses normal and anomalous examples for model adaptation during meta-training. Furthermore, we train ARCADe without meta-learning learning rates to investigate their impact when addressing a CAD problem (Question $3$). Finally, we analyze the distribution and properties of the learning rates meta-learned by ARCADe (Question $4$). \\

We evaluate ARCADe on three meta-learning benchmark datasets: Omniglot \cite{lake2011one}, MiniImageNet \cite{ravi2016optimization} and CIFAR-FS \cite{bertinetto2018meta}. Omniglot is composed of 20 instances of 1623 hand-written character classes from 50 different alphabets. The images have the size $28$x$28$ pixels. We use 25 alphabets for meta-training, 5 for meta-validation and 20 for meta-testing. MiniImageNet contains $100$ classes from ImageNet where each class includes $600$ images of size $84$x$84$x$3$. We use the official data split of $64$ classes for meta-training, $16$ for meta-validation and $20$ for meta-testing. CIFAR-FS was derived from CIFAR-100 by dividing its classes into $64$ classes for meta-training, $16$ for meta-validation and $20$ for meta-testing to make it suitable for meta-learning problems. Here, each class includes $600$ images of size $32$x$32$x$3$. The same data splits are used for ARCADe and the baselines. \\

To create meta-learning tasks for CAD, i.e. sequences of anomaly detection tasks as explained in Section \ref{problem-formulation}, we proceed as follows. First, we divide the classes available, e.g. the meta-training classes, into $L$ disjoint sets of classes, where $L$ is the task-sequence length. By building tasks using these sets we ensure that the tasks do not share any class. Subsequently, to create a task, one class from its set of classes is randomly chosen to be the normal class, i.e. its datapoints are labeled as non-anomalous, while the remaining classes are all labeled as anomalous. This ensures that the anomaly class has a higher variance than the normal class, which is usually the case in AD problems. Two disjoint sets of examples are then created from this task: a training set $T^{tr}$ containing only normal class examples and a class-balanced validation set $T^{val}$. The tasks are then concatenated in a random order into a task-sequence. This task-sequence creation procedure is adopted to create meta-training, meta-validation and meta-testing task-sequences for the three datasets. 
\\

For ARCADe as well as for the baselines we use the same 4-module architecture used in \cite{spigler2019meta} for continual learning. Each module includes a $3$x$3$ convolutional layer, a $2$x$2$ max-pooling layer, a batch-normalization \cite{ioffe2015batch} layer and a ReLU activation function. The 4 modules are followed by a linear layer and a sigmoid activation function. For omniglot, the convolutional layers include $64$ filters, while for MiniImageNet and CIFAR-FS they include $32$ filters. Since the meta-update of ARCADe requires backpropagating the gradients through all updates of all tasks, which is computationally expensive, we use a first-order approximation for our experiments. Hereby, the second-order terms of the derivatives are ignored, as done in \cite{spigler2019meta}.

\subsection{Results and discussion} \label{results}
\begin{figure}[h]
\caption{Retained accuracy on Omniglot}
\label{fig:arcade-vs-baselines-omn}
\begin{center}
\includegraphics[width=8cm]{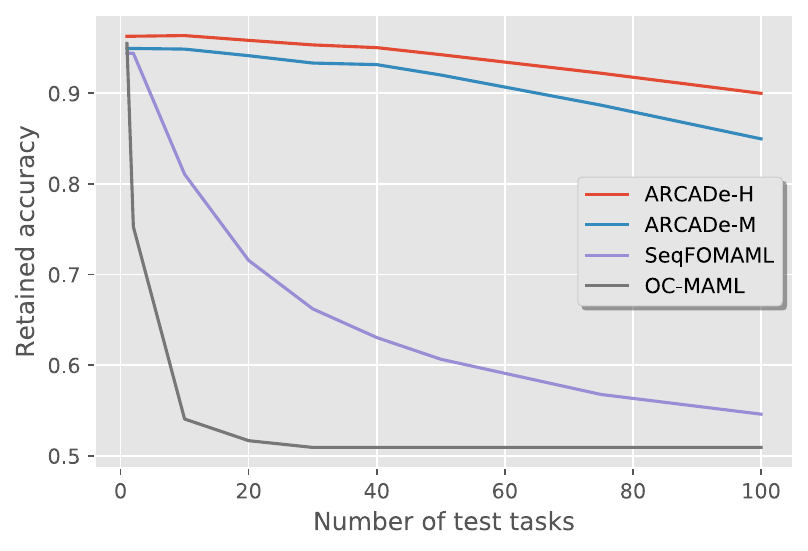}
\end{center}
\end{figure}

\begin{figure}[h]
\caption{Retained accuracy on MiniImageNet and CIFAR-FS}
\label{fig:arcade-vs-baselines-min-cifar}
\begin{center}
\includegraphics[width=8cm]{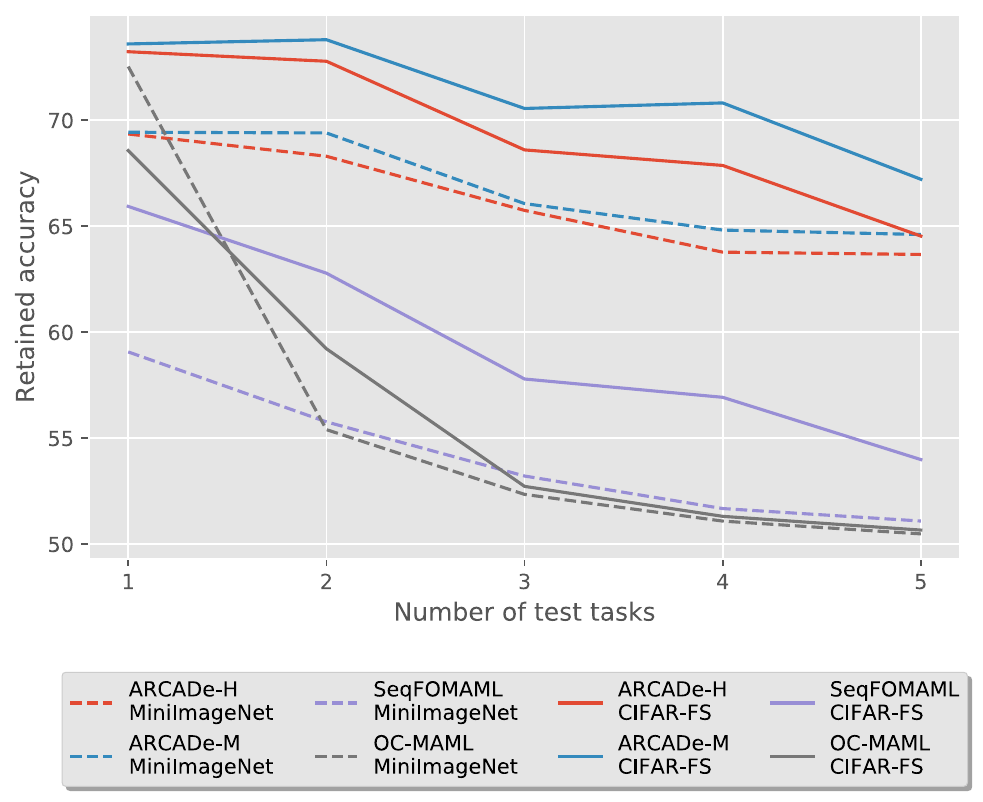}
\end{center}
\end{figure}

In this section we present and discuss the results of our experimental evaluation. Following previous continual learning works \cite{lopez2017gradient, riemer2018learning} we consider the final retained accuracy, i.e. the average of the accuracies of the final model on the validation sets of all test tasks, as our main metric. We use task-sequences composed of 10 tasks for meta-training on Omniglot and 5 tasks for meta-training on MiniImageNet and CIFAR-FS. For meta-testing task-sequence lengths between 1 and 100 are used for Omniglot and between 1 and 5 for the more challenging MiniImageNet and CIFAR-FS. During meta-training and meta-testing, each task is learned by performing only 3 gradient descent updates and using only 10 normal examples, across all datasets. This extends ARCADe's applicability to \emph{few-shot} CAD problems, i.e. CAD problems that exhibit extreme data scarcity. The performance of the two ARCADe variants and the baselines is shown in Figure \ref{fig:arcade-vs-baselines-omn} on Omniglot and in Figure \ref{fig:arcade-vs-baselines-min-cifar} on MiniImageNet and CIFAR-FS. For all datasets, we report the retained accuracy averaged over 500 task-sequences from the meta-testing set $D_{test}$.

We find that both ARCADe variants substantially outperform the baselines for all sequences that include more than one task on all three datasets. While the model initialization meta-learned by SeqFOMAML slows down catastrophic forgetting when adapted to class-balanced tasks \cite{spigler2019meta}, it fails at retaining a high accuracy in the CAD problem setting i.e. when adapted to a sequence of \emph{OCC} tasks. The quick decrease in retained accuracy suggests an important overfitting to the majority class. While OC-MAML yields a higher accuracy on the first task on MiniImageNet and Omniglot, it is not able to preserve this performance while learning the subsequent tasks in the sequence. In a CAD situation, the OC-MAML model quickly forgets the first task learned and collapses to a model that predicts only the majority class. We note that the lower performance of OC-MAML on the first task compared to the results reported in \cite{frikha2020fewshot} is due to the different evaluation setting in the CAD problem, where the identifiers and training sets of the learned tasks are not available at test time. OC-MAML uses the training set of the learned test task to overwrite the batch normalization statistics (mean and variance) before testing on the validation set.  \\

Surprisingly, we find that ARCADe can learn up to 100 OCC tasks sequentially on Omniglot, while losing only $6\%$ accuracy, even though it was trained with only 10-tasks sequences. We observe that ARCADe-H outperforms ARCADe-M on Omniglot, while ARCADe-M achieves higher retained accuracy on MiniImageNet and CIFAR-FS. Our explanation for this is that since MiniImageNet and CIFAR-FS have a higher variance in the input space, adapting the parameters of the feature extractor to the normal classes of the test tasks is beneficial. However, ARCADe-H can only adapt the parameters of the output layer, which results in a lower performance. The features meta-learned on the meta-training set of Omniglot, which includes by far more classes than the ones of MiniImageNet and CIFAR-FS, require less adaptation to perform well on the meta-testing set. \\


To assess the impact of meta-learning parameter specific learning rates, we evaluate ARCADe with constant learning rates, i.e. only parameter initializations are meta-learned. In Table \ref{tab:lrs-vs-no-lrs}, we present the results in terms of retained accuracy on test task-sequences with the same length as the ones used for meta-training. We find that additionally meta-learning learning rates boosts the performance of both ARCADe variants across all datasets. This validates our hypothesis that additionally meta-learning learning rates leads to a more effective inductive bias for the addressed CAD problem.

\begin{table}[h]
\caption{Retained test accuracies of ARCADe with and without meta-learning learning rates}
\label{tab:lrs-vs-no-lrs}
\vskip 0.1in
\begin{center}
\begin{tabular}{ l  l  l  l  }
\hline
Model $\backslash$ Dataset &Omniglot &CIFAR-FS &MIN   \\
\hline
ARCADe-M                      &$\bf96.1$           &$\bf68.1$    &$\bf64.5$  \\ 
ARCADe-M (constant $\alpha$)  &$95.7$      &$66.4$         &$63.1$   \\
ARCADe-H                      &$\bf96$           &$\bf67.8$    &$\bf64.1$ \\
ARCADe-H (constant $\alpha$)  &$95.6$      &$66.8$         &$63.0$  \\
\hline
\end{tabular}
\end{center}
\vskip -0.1in
\end{table}

\begin{figure}[h]
\caption{Layer-wise mean and percentage of positive learning rates meta-learned by ARCADe-M}
\label{fig:metalearned-lrs}
\begin{center}
\includegraphics[width=8cm]{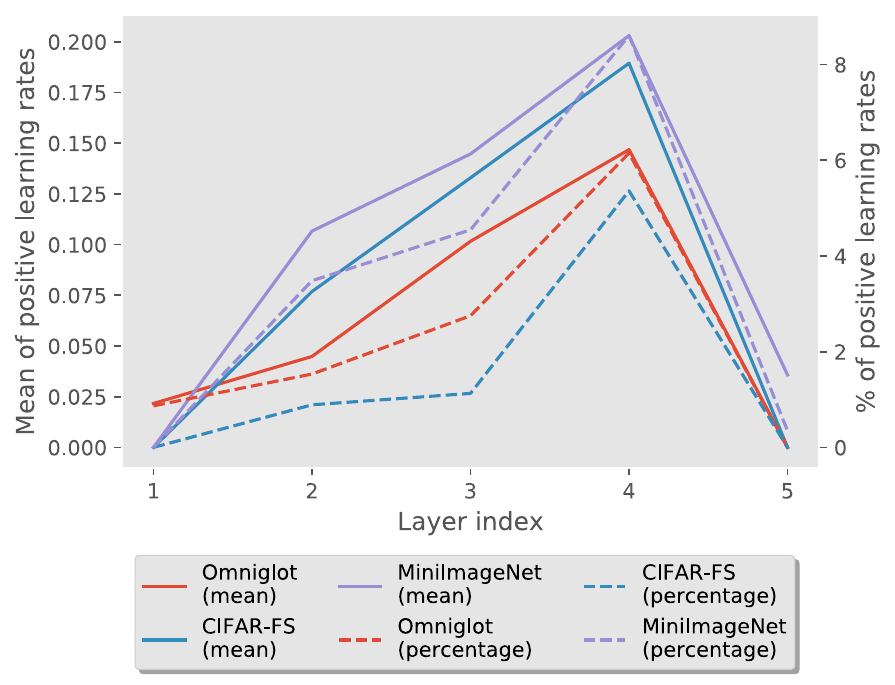}
\end{center}
\end{figure}

Finally, we would like to investigate the characteristics of the meta-learned learning rates in order to gain a deeper insight into the learning strategy to which ARCADe-M converges. As mentioned in Section \ref{approach}, we clip the learning rates between 0 and 1. Thus, only positive learning rates are active. We measure the percentage and mean of the positive (active) learning rate per neural network layer and present them in Figure \ref{fig:metalearned-lrs}.

The following observations and interpretations hold for all three datasets. We find that, for all layers, the majority of the learning rates are chosen to be not active, i.e. they converge to negative values. This suggests that most parameters are task-agnostic and can be used as-is independently of the task-sequence to be learned. As we progress in the layers of the embedding network (CNN), the percentage of active learning rates increases to reach its maximum at the last convolutional layer. This means that, while the basic features (layer 1) can be reused without adaptation across tasks, the more sophisticated features that are used for classification (layer 4) have to be task-specifically adapted. Moreover, ARCADe-M freezes almost all the parameters in the linear output layer (less than $1\%$ of the parameters are updated). This suggests that, each time it learns a new task, ARCADe-M does not update its normal class decision boundary to include the embeddings of the normal class examples of this new task, but rather changes the embedding of the latter to fit inside a frozen decision boundary. We hypothesize that ARCADe-M does this since the output layer is more prone to the CAD challenges, i.e. overfitting to the majority class and catastrophic forgetting. \\

Furthermore, we analyze the means of the active learning rates and find a similar trend across the layers. In fact, the few active learning rates in the first and last layer have substantially lower values than those of the other convolutional layers, especially layer 4. This shows that, during adaptation, bigger update steps are performed on the last layer of the embedding network than on the output layer, which backs our previous interpretation of the ARCADe-M's learning strategy. On the other hand, ARCADe-H cannot update the parameters of the embedding network by design, and learns therefore how to adapt the parameter of the output layer. Analyzing the parameter-specific learning rates meta-learned by ARCADe-H shows also that some parameters are chosen to be task-agnostic (due to negative learning rates), while other are chosen to be task-specific. This further explains the performance increase of ARCADe-H when additionally meta-learning learning rates (Table \ref{tab:lrs-vs-no-lrs}).

\section{Conclusion}
In this work we addressed the novel and challenging problem of Continual Anomaly Detection (CAD). After formulating this learning scenario as a meta-learning problem, we proposed \emph{A Rapid Continual Anomaly Detector (ARCADe)} to serve as a first and strong baseline in this research context. On the Omniglot dataset, our meta-learning approach enables sequentially learning up to 100 anomaly detection tasks using only examples from their normal (majority) class, with minimal forgetting and overfitting to the majority class. Our method substantially outperformed continual learning and anomaly detection baselines on three datasets.  

\bibliographystyle{IEEEtran}
\bibliography{references}

\begin{thebibliography}{10}
\providecommand{\url}[1]{#1}
\csname url@samestyle\endcsname
\providecommand{\newblock}{\relax}
\providecommand{\bibinfo}[2]{#2}
\providecommand{\BIBentrySTDinterwordspacing}{\spaceskip=0pt\relax}
\providecommand{\BIBentryALTinterwordstretchfactor}{4}
\providecommand{\BIBentryALTinterwordspacing}{\spaceskip=\fontdimen2\font plus
\BIBentryALTinterwordstretchfactor\fontdimen3\font minus
  \fontdimen4\font\relax}
\providecommand{\BIBforeignlanguage}[2]{{%
\expandafter\ifx\csname l@#1\endcsname\relax
\typeout{** WARNING: IEEEtran.bst: No hyphenation pattern has been}%
\typeout{** loaded for the language `#1'. Using the pattern for}%
\typeout{** the default language instead.}%
\else
\language=\csname l@#1\endcsname
\fi
#2}}
\providecommand{\BIBdecl}{\relax}
\BIBdecl

\bibitem{french1999catastrophic}
R.~M. French, ``Catastrophic forgetting in connectionist networks,''
  \emph{Trends in cognitive sciences}, 1999.

\bibitem{goodrich2015neuron}
B.~F. Goodrich, ``Neuron clustering for mitigating catastrophic forgetting in
  supervised and reinforcement learning,'' 2015.

\bibitem{kirkpatrick2017overcoming}
J.~Kirkpatrick, R.~Pascanu, N.~Rabinowitz, J.~Veness, G.~Desjardins, A.~A.
  Rusu, K.~Milan, J.~Quan, T.~Ramalho, A.~Grabska-Barwinska \emph{et~al.},
  ``Overcoming catastrophic forgetting in neural networks,'' \emph{Proceedings
  of the national academy of sciences}, 2017.

\bibitem{caruana1997multitask}
R.~Caruana, ``Multitask learning,'' \emph{Machine learning}, 1997.

\bibitem{zenke2017continual}
F.~Zenke, B.~Poole, and S.~Ganguli, ``Continual learning through synaptic
  intelligence,'' \emph{Proceedings of machine learning research}, 2017.

\bibitem{lee2017overcoming}
S.-W. Lee, J.-H. Kim, J.~Jun, J.-W. Ha, and B.-T. Zhang, ``Overcoming
  catastrophic forgetting by incremental moment matching,'' in \emph{Advances
  in neural information processing systems}, 2017.

\bibitem{lopez2017gradient}
D.~Lopez-Paz and M.~Ranzato, ``Gradient episodic memory for continual
  learning,'' in \emph{Advances in neural information processing systems},
  2017.

\bibitem{chaudhry2018riemannian}
A.~Chaudhry, P.~K. Dokania, T.~Ajanthan, and P.~H. Torr, ``Riemannian walk for
  incremental learning: Understanding forgetting and intransigence,'' in
  \emph{Proceedings of the European Conference on Computer Vision (ECCV)},
  2018.

\bibitem{riemer2018learning}
M.~Riemer, I.~Cases, R.~Ajemian, M.~Liu, I.~Rish, Y.~Tu, and G.~Tesauro,
  ``Learning to learn without forgetting by maximizing transfer and minimizing
  interference,'' 2018.

\bibitem{javed2019meta}
K.~Javed and M.~White, ``Meta-learning representations for continual
  learning,'' in \emph{Advances in Neural Information Processing Systems},
  2019.

\bibitem{beaulieu2020learning}
S.~Beaulieu, L.~Frati, T.~Miconi, J.~Lehman, K.~O. Stanley, J.~Clune, and
  N.~Cheney, ``Learning to continually learn,'' 2020.

\bibitem{chandola2009anomaly}
V.~Chandola, A.~Banerjee, and V.~Kumar, ``Anomaly detection: A survey,''
  \emph{ACM computing surveys (CSUR)}, 2009.

\bibitem{aljundi2019task}
R.~Aljundi, K.~Kelchtermans, and T.~Tuytelaars, ``Task-free continual
  learning,'' in \emph{Proceedings of the IEEE Conference on Computer Vision
  and Pattern Recognition}, 2019.

\bibitem{wang2018systematic}
S.~Wang, L.~L. Minku, and X.~Yao, ``A systematic study of online class
  imbalance learning with concept drift,'' \emph{IEEE transactions on neural
  networks and learning systems}, 2018.

\bibitem{moya1993one}
M.~M. Moya, M.~W. Koch, and L.~D. Hostetler, ``One-class classifier networks
  for target recognition applications,'' \emph{NASA STI/Recon Technical Report
  N}, 1993.

\bibitem{khan2014one}
S.~S. Khan and M.~G. Madden, ``One-class classification: taxonomy of study and
  review of techniques,'' \emph{The Knowledge Engineering Review}, 2014.

\bibitem{schmidhuber1987evolutionary}
J.~Schmidhuber, ``Evolutionary principles in self-referential learning, or on
  learning how to learn: the meta-meta-... hook,'' Ph.D. dissertation,
  Technische Universit{\"a}t M{\"u}nchen, 1987.

\bibitem{spigler2019meta}
G.~Spigler, ``Meta-learnt priors slow down catastrophic forgetting in neural
  networks,'' 2019.

\bibitem{zhang2019variational}
M.~Zhang, T.~Wang, J.~H. Lim, G.~Kreiman, and J.~Feng, ``Variational prototype
  replays for continual learning,'' 2019.

\bibitem{finn2017model}
C.~Finn, P.~Abbeel, and S.~Levine, ``Model-agnostic meta-learning for fast
  adaptation of deep networks,'' in \emph{Proceedings of the 34th International
  Conference on Machine Learning}, 2017.

\bibitem{vinyals2016matching}
O.~Vinyals, C.~Blundell, T.~Lillicrap, K.~Kavukcuoglu, and D.~Wierstra,
  ``Matching networks for one shot learning,'' 2016.

\bibitem{wang2019few}
Y.~Wang and Q.~Yao, ``Few-shot learning: A survey,'' 2019.

\bibitem{li2017meta}
Z.~Li, F.~Zhou, F.~Chen, and H.~Li, ``Meta-sgd: Learning to learn quickly for
  few-shot learning,'' 2017.

\bibitem{lee2019meta}
K.~Lee, S.~Maji, A.~Ravichandran, and S.~Soatto, ``Meta-learning with
  differentiable convex optimization,'' in \emph{Proceedings of the IEEE
  Conference on Computer Vision and Pattern Recognition}, 2019.

\bibitem{raghu2019rapid}
A.~Raghu, M.~Raghu, S.~Bengio, and O.~Vinyals, ``Rapid learning or feature
  reuse? towards understanding the effectiveness of maml,'' 2019.

\bibitem{frikha2020fewshot}
A.~Frikha, D.~Krompaß, H.-G. Köpken, and V.~Tresp, ``Few-shot one-class
  classification via meta-learning,'' 2020.

\bibitem{yin2018adversarial}
C.~Yin, J.~Tang, Z.~Xu, and Y.~Wang, ``Adversarial meta-learning,'' 2018.

\bibitem{rusu2016progressive}
A.~A. Rusu, N.~C. Rabinowitz, G.~Desjardins, H.~Soyer, J.~Kirkpatrick,
  K.~Kavukcuoglu, R.~Pascanu, and R.~Hadsell, ``Progressive neural networks,''
  2016.

\bibitem{schaul2015prioritized}
T.~Schaul, J.~Quan, I.~Antonoglou, and D.~Silver, ``Prioritized experience
  replay,'' 2015.

\bibitem{he2019task}
X.~He, J.~Sygnowski, A.~Galashov, A.~A. Rusu, Y.~W. Teh, and R.~Pascanu, ``Task
  agnostic continual learning via meta learning,'' 2019.

\bibitem{nichol2018reptile}
A.~Nichol and J.~Schulman, ``Reptile: a scalable metalearning algorithm,''
  2018.

\bibitem{snell2017prototypical}
J.~Snell, K.~Swersky, and R.~Zemel, ``Prototypical networks for few-shot
  learning,'' in \emph{Advances in Neural Information Processing Systems},
  2017.

\bibitem{scholkopf2001estimating}
B.~Sch{\"o}lkopf, J.~C. Platt, J.~Shawe-Taylor, A.~J. Smola, and R.~C.
  Williamson, ``Estimating the support of a high-dimensional distribution,''
  \emph{Neural computation}, 2001.

\bibitem{tax2004support}
D.~M. Tax and R.~P. Duin, ``Support vector data description,'' \emph{Machine
  learning}, 2004.

\bibitem{Xu_2015}
D.~Xu, E.~Ricci, Y.~Yan, J.~Song, and N.~Sebe, ``Learning deep representations
  of appearance and motion for anomalous event detection,'' \emph{Procedings of
  the British Machine Vision Conference 2015}, 2015.

\bibitem{andrews2016transfer}
J.~T. Andrews, T.~Tanay, E.~J. Morton, and L.~D. Griffin, ``Transfer
  representation-learning for anomaly detection.''\hskip 1em plus 0.5em minus
  0.4em\relax ICML, 2016.

\bibitem{erfani2016high}
S.~M. Erfani, S.~Rajasegarar, S.~Karunasekera, and C.~Leckie,
  ``High-dimensional and large-scale anomaly detection using a linear one-class
  svm with deep learning,'' \emph{Pattern Recognition}, 2016.

\bibitem{ruff2018deep}
L.~Ruff, R.~Vandermeulen, N.~Goernitz, L.~Deecke, S.~A. Siddiqui, A.~Binder,
  E.~M{\"u}ller, and M.~Kloft, ``Deep one-class classification,'' in
  \emph{International Conference on Machine Learning}, 2018.

\bibitem{hinton2006reducing}
G.~E. Hinton and R.~R. Salakhutdinov, ``Reducing the dimensionality of data
  with neural networks,'' \emph{science}, 2006.

\bibitem{hawkins2002outlier}
S.~Hawkins, H.~He, G.~Williams, and R.~Baxter, ``Outlier detection using
  replicator neural networks,'' in \emph{International Conference on Data
  Warehousing and Knowledge Discovery}.\hskip 1em plus 0.5em minus 0.4em\relax
  Springer, 2002.

\bibitem{an2015variational}
J.~An and S.~Cho, ``Variational autoencoder based anomaly detection using
  reconstruction probability,'' \emph{Special Lecture on IE}, 2015.

\bibitem{chen2017outlier}
J.~Chen, S.~Sathe, C.~Aggarwal, and D.~Turaga, ``Outlier detection with
  autoencoder ensembles,'' in \emph{Proceedings of the 2017 SIAM International
  Conference on Data Mining}.\hskip 1em plus 0.5em minus 0.4em\relax SIAM,
  2017.

\bibitem{goodfellow2014generative}
I.~Goodfellow, J.~Pouget-Abadie, M.~Mirza, B.~Xu, D.~Warde-Farley, S.~Ozair,
  A.~Courville, and Y.~Bengio, ``Generative adversarial nets,'' in
  \emph{Advances in neural information processing systems}, 2014.

\bibitem{Ravanbakhsh_2017}
M.~Ravanbakhsh, M.~Nabi, E.~Sangineto, L.~Marcenaro, C.~Regazzoni, and N.~Sebe,
  ``Abnormal event detection in videos using generative adversarial nets,''
  \emph{2017 IEEE International Conference on Image Processing (ICIP)}, 2017.

\bibitem{schlegl2017unsupervised}
T.~Schlegl, P.~Seeb{\"o}ck, S.~M. Waldstein, U.~Schmidt-Erfurth, and G.~Langs,
  ``Unsupervised anomaly detection with generative adversarial networks to
  guide marker discovery,'' in \emph{International Conference on Information
  Processing in Medical Imaging}.

\bibitem{Sabokrou_2018}
M.~Sabokrou, M.~Khalooei, M.~Fathy, and E.~Adeli, ``Adversarially learned
  one-class classifier for novelty detection,'' \emph{2018 IEEE/CVF Conference
  on Computer Vision and Pattern Recognition}, 2018.

\bibitem{finn2017metalearning}
C.~Finn and S.~Levine, ``Meta-learning and universality: Deep representations
  and gradient descent can approximate any learning algorithm,'' 2017.

\bibitem{lake2011one}
B.~Lake, R.~Salakhutdinov, J.~Gross, and J.~Tenenbaum, ``One shot learning of
  simple visual concepts,'' in \emph{Proceedings of the annual meeting of the
  cognitive science society}, 2011.

\bibitem{ravi2016optimization}
S.~Ravi and H.~Larochelle, ``Optimization as a model for few-shot learning,''
  2016.

\bibitem{bertinetto2018meta}
L.~Bertinetto, J.~F. Henriques, P.~H. Torr, and A.~Vedaldi, ``Meta-learning
  with differentiable closed-form solvers,'' 2018.

\bibitem{ioffe2015batch}
S.~Ioffe and C.~Szegedy, ``Batch normalization: Accelerating deep network
  training by reducing internal covariate shift,'' \emph{arXiv preprint
  arXiv:1502.03167}, 2015.

\end{thebibliography}

\end{document}